\documentstyle[aaai]{article}
\input psfig
\begin{document}

\title{\begin{small} In AAAI-93: Proceedings of the Eleventh National 
Conference on Artificial Intelligence, July 11-15, 1993, Washington,
DC, pp 380-385. \end{small} \\ Having Your Cake and Eating It Too:
Autonomy and Interaction in a Model of Sentence Processing}

\author{Kurt P. Eiselt$^{*}$ \\
{\normalsize College of Computing} \\
{\normalsize Georgia Institute of Technology} \\
{\normalsize Atlanta, Georgia 30332-0280} \\
{\normalsize eiselt@cc.gatech.edu} \And
Kavi Mahesh\thanks{During the course of this work, these authors were 
supported in part by a 
research grant from Northern Telecom.} \\
{\normalsize College of Computing} \\
{\normalsize Georgia Institute of Technology} \\
{\normalsize Atlanta, Georgia 30332-0280} \\
{\normalsize mahesh@cc.gatech.edu} \And
Jennifer K. Holbrook \\
{\normalsize Department of Psychology} \\
{\normalsize Albion College} \\
{\normalsize Albion, Michigan 49224} \\
{\normalsize jen@cedar.cic.net}}

\maketitle

\begin{abstract}
Is the human language understander a collection of modular processes
operating with relative autonomy, or is it a single integrated
process?  This ongoing debate has polarized the language processing
community, with two fundamentally different types of model posited,
and with each camp concluding that the other is wrong. One camp puts
forth a model with separate processors and distinct knowledge sources
to explain one body of data, and the other proposes a model with a
single processor and a homogeneous, monolithic knowledge source to
explain the other body of data.  In this paper we argue that a hybrid
approach which combines a unified processor with separate knowledge
sources provides an explanation of both bodies of data, and we
demonstrate the feasibility of this approach with the computational
model called COMPERE. We believe that this approach brings the
language processing community significantly closer to offering
human-like language processing systems.
\end{abstract}
						
\section{The Big Questions}

Years of research by linguists, psychologists, and artificial
intelligence specialists have provided significant insight into the
workings of the human language processor.  Still, fundamental
questions remain unanswered.  In particular, the debate over modular
processing versus integrated processing rages on, and experimental
data and computational models exist to support both positions.
Furthermore, if the integrated processing position is correct, just
what exactly is integrated?  And if the modular position is the right
one, what are the different modules?  Do they interact, and if so, to
what extent and when?  Or are those modules entirely autonomous?

Wrestling with these questions induces considerable frustration in
researchers.  This frustration stems not only from the research
community's apparent inability to answer them satisfactorily, but also
from the overwhelming importance of the answers themselves---these
answers, once uncovered, undoubtedly will impact thinking in all areas
of artificial intelligence and cognitive science research, including
visual processing, reasoning, and problem solving, to name just a few.
In this paper, we intend to provide the reader with answers to some of
these questions---answers based on nearly ten years of our own
interdisciplinary research in sentence processing, and built upon the
work of many others who went before us.  In brief, we propose a model
of language understanding (or, more specifically, sentence processing)
in which all linguistic processing is performed by a single unified
process, but the different types of linguistic knowledge necessary for
processing are separate and distinct.  This model accounts for
conflicting experimental data, some of which suggests an autonomous,
modular approach to language processing, and some of which indicates
an integrated approach.  Because it is a closer fit to the
experimental data than any model which has gone before, this model
consequently points the way to more human-like performance from
language processing systems.

\section{Background}

Our new model of sentence processing has its roots in work begun
nearly ten years ago.  That research effort started as an attempt to
explain how the human language understander selected the most
context-appropriate meaning of an ambiguous word, and then was able to
correct both the choice of word meaning and the surrounding sentence
interpretation, without reprocessing the input, when later processing
showed that the initial choice of word meaning was erroneous.

The resulting computational model, ATLAST (Eiselt, 1987; Eiselt,
1989), resolved word sense ambiguities by activating multiple word
meanings in parallel, selecting the meaning which matched the previous
context, and deactivating but retaining the unchosen meanings for as
long as resources were available for retaining them.  If later context
proved the initial decision to be incorrect, the retained meanings
could be reactivated without reaccessing the lexicon or reprocessing
the text.  ATLAST proved to have great psychological validity for
lexical processing---its use of multiple access was well grounded in
the psychological literature (e.g., Seidenberg, Tanenhaus, Leiman, \&
Bienkowski, 1982), and, more importantly, it made psychological
predictions about the retention of unselected meanings that were
experimentally validated (Eiselt \& Holbrook, 1991; Holbrook, 1989).
ATLAST provided an architecture of sentence processing which was also
used to explain recovery from erroneous decisions in making pragmatic
inferences as well as explaining individual differences in pragmatic
inferences (Eiselt, 1989; cf. Granger, Eiselt, \& Holbrook, 1983).

Error recovery in semantic processing had occasionally aroused the
attention of researchers in conceptually-based natural language
understanding, but the questions that arose were usually dismissed as
unimportant or something which could be resolved as an afterthought
(Birnbaum \& Selfridge, 1981; Lebowitz, 1980; Lytinen, 1984).  These
researchers were content to assume that the first inference decision
made was the correct one.  Meanwhile, other researchers investigating
syntactically-based approaches had long since concluded that the means
by which erroneous syntactic decisions were accommodated had a
dramatic impact on the architecture of the syntactic processor being
proposed.  For example, the backtracking models embodied the theory
that only a single syntactic interpretation need be maintained at any
given time, so long as the processor could keep track of its
decisions, undo them when an erroneous decision was discovered, and
then reinterpret the input (e.g., Woods, 1973).  The lookahead parsers
tried to sidestep the problems inherent in backtracking by postponing
any decision until enough input had been processed to guarantee a
correct decision, thereby avoiding erroneous decisions to some extent
(e.g., Marcus, 1980).  Another approach to avoiding erroneous
decisions was offered by parallel parsers which maintained all
plausible syntactic interpretations at the same time (Kurtzman, 1985).
ATLAST, however, was a model of semantic processing and did not
address the issue of recovery from erroneous syntactic decisions, nor
did it substantially address the issue of syntactic processing at all.

Recently, Stowe (1991) presented experimental evidence showing that in
dealing with syntactic ambiguity, the sentence processor accesses all
possible syntactic structures simultaneously and, if the structure
preferred for syntactic reasons conflicts with the structure favored
by the current semantic bias, the competing structures are maintained
and the decision is delayed.  Furthermore, the work suggests an
interaction of the various knowledge types, as in some cases semantic
information influences structure assignment or triggers reactivation
of unselected structures.  This model of limited delayed decision in
syntactic ambiguity resolution had much in common with the ATLAST
model of semantic ambiguity resolution.  Both models proposed an early
commitment where possible.  Both models had the capability to pursue
multiple interpretations in parallel when ambiguity made it necessary.
Both models explained error recovery as an operation of switching to
another interpretation maintained in parallel by the sentence
processor.  Finally, both models made decisions by integrating the
preferences from syntax and semantics.

One explanation for this high degree of similarity between the
syntactic and semantic error recovery mechanisms is that there are two
separate processors, one for syntax and one for semantics, each with
its corresponding source of linguistic knowledge, and each doing
exactly the same thing.  A more economical explanation, however, is
that there is only one process which deals with syntactic and semantic
information in the same manner.  We have chosen to explore the latter
explanation, as others have done, but we have also chosen to maintain
the separate knowledge sources for reasons which will be explained
below.  (See also Holbrook, Eiselt, \& Mahesh, 1992.)

\section{Overview of COMPERE}

Our new model of sentence processing, called COMPERE (Cognitive Model
of Parsing and Error Recovery), consists of a single unified process
operating on independent sources of syntactic and semantic knowledge.
This is made possible by a uniform representation of both types of
knowledge.  The unified process applies the same operations to the
different types of knowledge, and has a single control structure which
performs the operations on syntactic and semantic knowledge in tandem.
This permits a rich interaction between the two sources of knowledge,
both through transfer of control and through a shared representation
of the interpretations of the input text being built by the unified
process.

An advantage of representing the different kinds of knowledge in the
same form is that the boundaries between the different types of
knowledge can be ill-defined.  Often it is difficult to classify a
piece of knowledge as belonging to a particular class such as
syntactic or semantic.  With a uniform representation, such knowledge
lies in between and can be treated as belonging to either class.

Syntactic and semantic knowledge are represented in separate networks
in which each node is a structured representation of all the
information pertaining to a syntactic or semantic category or concept.
A link, represented as a slot-filler pair in the node, specifies a
parent category or concept of which the node can be a part, together
with the conditions under which it can be bound to the parent, and the
expectations that are certain to be fulfilled should the node be bound
to the parent.  In addition, nodes in either network are linked to
corresponding nodes in the other network so that the unified process
can build on-line interpretations of the input sentence in which each
syntactic unit has a corresponding representation of its thematic role
and its meaning. In addition, there is a lexicon as
well as certain other minor heuristic and control knowledge that is
part of the process.  (COMPERE's architecture and knowledge
representation are displayed graphically in Figures 1 and 2.)

The unified process is a bottom-up, early-commitment parsing mechanism
integrated with top-down guidance through expectations. The operators
and the control structure that constitute the unified process are
described briefly in the algorithm shown in Figure 3.

\begin{figure}[htbp]
\begin{center}
\ \psfig{figure=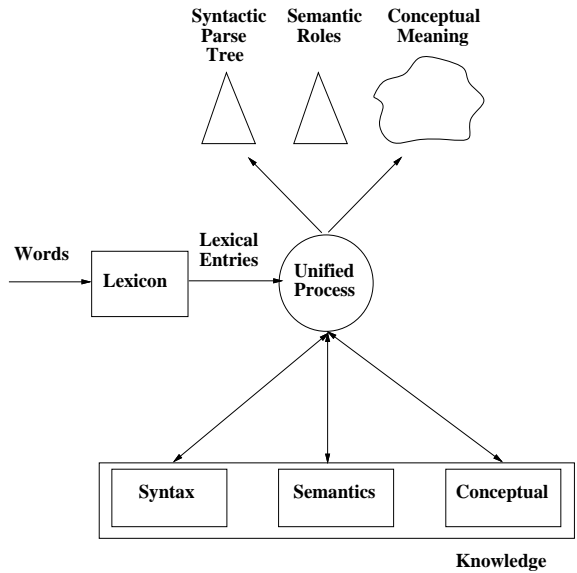,width=3.0in,height=3.0in}
\caption{Architecture of COMPERE.}
\end{center}
\end{figure}

\begin{figure}[htbp]
\begin{center}
\ \psfig{figure=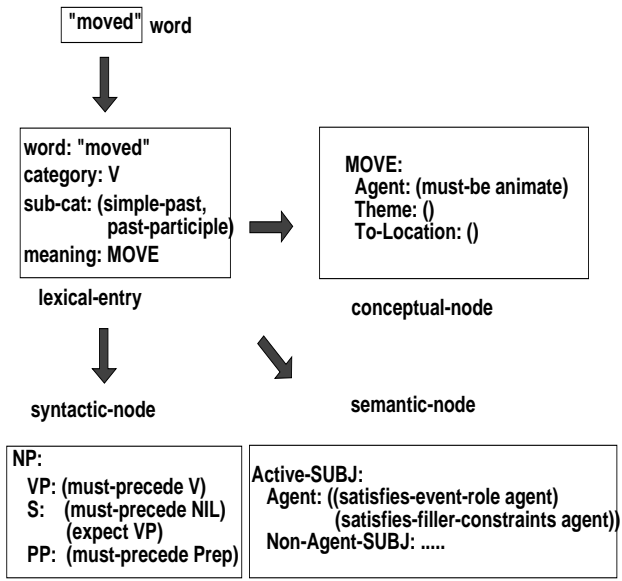,width=3.2in,height=3.0in}
\caption{Knowledge Representation in COMPERE.\footnotemark\ }
\end{center}
\end{figure}

\begin{figure}[htbp]
\noindent{}------------------------------ \\
1. Access lexical entries of next word. \\
2. Create instance nodes for syntactic category, meaning, and
   (primitive) thematic role. \\
3. Compute feasible bindings to parents for syntactic instance node
   and role instance node.  (This operation checks any conditions to
   be satisfied to make the binding feasible; it also takes existing
   expectations into account.) \\
4. Rank syntactic and semantic feasible bindings by their respective
   preference criteria. \\ Combine feasible bindings and select the
   most preferred binding. \\
5. Make the binding by creating parent node instances and appropriate 
   links, and generating any expectations. Create links between 
   corresponding instances in syntax and their thematic roles and 
   meanings. \\
6. Retain alternative bindings for possible error recovery. \\
7. If there is no feasible binding for a node, explore previously 
   retained alternatives to recover from errors. \\
8. Continue to bind the parent nodes to nodes further up as far as 
   possible (such as until the S node in syntax or the Event node in 
   semantics). \\
------------------------------
\caption{Unified Process: Algorithm.}
\end{figure}

The COMPERE prototype has been implemented in Common LISP on a
Symbolics LISP Machine.  At this time, its unified process can perform
on-line interpretations of its input, and can recover from erroneous
syntactic decisions when necessary.  COMPERE is able to process
relatively complex syntactic structures, including relative clauses,
and can resolve the associated structural ambiguities.

\footnotetext{The arrows
in Figure 2 simply indicate which types of knowledge point to which
other types; they do not mean that the specific nodes shown point to
the other nodes.}

\section{Autonomy and interaction effects from one process}

COMPERE is able to exhibit seemingly modular processing behavior that
matches the results of experiments showing the autonomy of different
levels of language processing (e.g., Forster, 1979; Frazier, 1987).
It is also able to display seemingly integrated behavior that matches
the results of experiments showing semantic influences on syntactic
structure assignment (e.g., Crain \& Steedman, 1985; Tyler \&
Marslen-Wilson, 1977).  For example, consider the processing of the
following sentence:
\smallskip

\noindent{\bf (1)} {\em The bugs moved into the new lounge were found quickly.}

\smallskip
This sentence has a lexical semantic ambiguity at the subject noun
{\em bugs} that could mean either insects or electronic
microphones. In addition, it is also syntactically ambiguous locally
at the verb {\em moved} since there is no distinction between its
past-tense form and its past-participle form.  In the simple past
reading of {\em moved}, it would be the main verb with the
corresponding interpretation that ``the bugs moved themselves into the
new lounge.''  On the other hand, if {\em moved} is read as a verb in
its past-participle form, it would be the verb in a reduced relative
clause corresponding to the meaning ``the bugs which were moved by
somebody else into the new lounge....'' Parse trees for the two
structural interpretations and the corresponding thematic-role
assignments are shown in Figures 4 and 5.\footnote{For simplicity,
these figures show the parse trees and the thematic roles separate
from each other.  In COMPERE's actual output, the parse trees and
thematic roles are interlinked.} \\

\begin{figure}[htbp]
\begin{center}
\ \psfig{figure=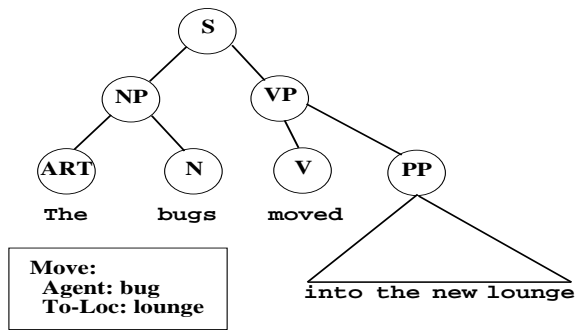,width=3.0in,height=1.7in}
\caption{Garden Path: Main-Clause Interpretation.}
\end{center}
\end{figure}

\smallskip
\noindent{\bf Null Context:} When sentence (1) is presented to COMPERE in a 
null semantic context, one where there is no bias for either meaning
of the noun {\em bugs,} COMPERE reads ahead without resolving the
lexical ambiguity at the word {\em bugs.}  When it encounters the
structural ambiguity at the verb {\em moved,} COMPERE does not have
the necessary information to decide which of the two structures in
Figures 4 and 5 is the appropriate one to pursue.

However, COMPERE has a syntactic preference for the main-verb
interpretation over the relative clause one.  Though this preference
can be explained by the minimal attachment principle (Frazier, 1987),
COMPERE offers a more general explanation.  Extrapolating from Stowe's
model, we have endowed COMPERE with the pervasive goal of completing
an incomplete item at any level of processing.  In syntactic
processing, it has a goal to complete the syntactic structure of a
unit such as a phrase, clause, or a sentence.  COMPERE prefers the
alternative which helps complete the current structure (called the
Syntactic Default) over one that adds an optional constituent leaving
the incompleteness intact.  For instance, in (1), a VP is required to
complete the sentence after seeing {\em The bugs.}  Since the
main-clause interpretation helps complete this requirement and the
relative-clause interpretation does not, the main-clause structure
gets selected.  In other words, COMPERE would rather use the verb to
begin the VP that is required to complete the sentence structure than
treat it as the verb in a reduced relative clause which would leave
the expectation of the VP unsatisfied.  This behavior is the same as
the one explained by the ``first analysis'' models of Frazier and
colleagues (Frazier, 1987) using a minimal-attachment
preference. COMPERE can produce this behavior by applying structural
preferences independently since it maintains separate representations
of syntactic and semantic knowledge.

As a consequence of choosing the main-clause interpretation, the
lexical ambiguity is also resolved. The electronic bug meaning is now
ruled out since there is a selectional restriction on the verb {\em
moved} that is not satisfied by electronic bugs (namely, they cannot
move by themselves).\footnote{COMPERE's program does not resolve
lexical semantic ambiguities at this time. We are currently rectifying
this by incorporating lexical ambiguity resolution strategies from our
earlier model ATLAST (Eiselt, 1989) in COMPERE.} \\

\begin{figure}[htbp]
\begin{center}
\ \psfig{figure=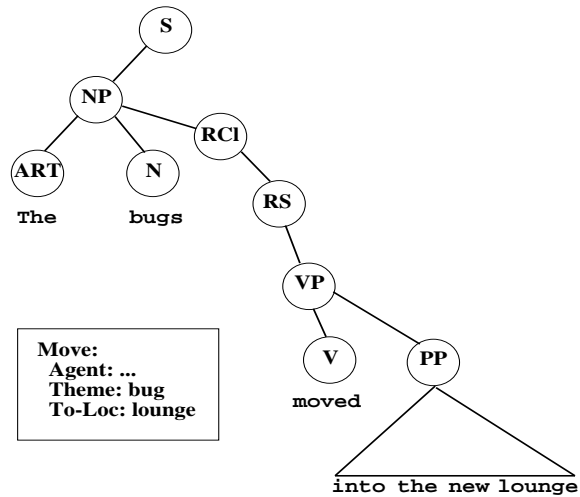,width=3.0in,height=2.6in}
\caption{Garden-Path: Reduced Relative Clause.}
\end{center}
\end{figure}
\smallskip
Thus, until seeing the word {\em were,} the verb {\em moved} is
treated as the main verb since it satisfies the expectation of a VP
that is required to complete the sentence.  However, at this point,
the structure is incompatible with the remaining input.  COMPERE
recognizes the error and now tries the alternative of attaching the VP
as a reduced relative clause so that there is still a place for a main
verb.  This results in a garden-path effect upon reading this
sentence.  That is, the sentence processor is led up a garden path and
has to backtrack when later information shows that it was the wrong
path to take.  This behavior is not influenced by semantic or
conceptual preferences and can be perceived as a modular
behavior. COMPERE's error recovery method was first developed in the
ATLAST model (Eiselt, 1987). It was also experimentally validated
(Eiselt \& Holbrook, 1991).

As a consequence of switching to the new syntactic interpretation,
COMPERE makes corresponding changes to thematic role assignments and
also ``unresolves'' the lexical ambiguity.  There is no longer any
reason to eliminate the electronic bug meaning since either kind of
bugs can be moved by others.

\noindent{\bf Semantically Biasing Context:} Now consider sentence (1) 
in a semantically biasing context such as the one in (2).\footnote{At
present, COMPERE is not capable of using context effects in its
ambiguity resolution process. However, its architecture supports the
inclusion of such effects and we are working on providing context
information to the unified process.} \\

\noindent{\bf (2)} {\em The Americans built a new wing to the embassy. The 
Russian spies quickly transferred the microphones to the new wing. The
bugs moved into the new lounge were found quickly.} \\

\smallskip
\noindent{}The semantic context in (2) resolves the lexical ambiguity by choosing
the electronic bug meaning. This decision helps COMPERE resolve the
structural ambiguity at the verb {\em moved.} Using its conceptual
knowledge, represented as a selectional restriction, that only animate
agents can move by themselves, COMPERE decides that {\em moved} cannot
be a main verb and goes directly to the reduced relative clause
interpretation (Fig. 5), thereby avoiding the garden path.  This shows
how the same unified process that previously exhibited modular
processing behavior can also produce interactive processing behavior
when semantic information is available. Syntax and semantics interact
in COMPERE to help resolve ambiguities in each other.

COMPERE can also use independent syntactic preferences in other types
of sentences such as those with prepositional attachment ambiguities.
The COMPERE prototype thus demonstrates that the range of behaviors
that the interactive models account for (Crain \& Steedman, 1985;
Tyler \& Marslen-Wilson, 1977), and the behaviors that the ``first
analysis'' models account for (Frazier, 1987), can be explained by a
unified model with a single processor operating on multiple
independent sources of knowledge.

\section{Comparative evaluation}

There is certainly nothing unique about a unified process model of
language understanding---the integrated processing hypothesis has been
visited and revisited many times, for good reason, and with
significant results (e.g., Jurafsky, 1992; Lebowitz, 1980; Riesbeck \&
Martin, 1986).  Yet each of these models labors under the assumption
that the integration of processing necessarily goes hand in hand with
the integration of the knowledge sources.  While this design decision
may make construction of the corresponding computational model easier,
it also makes the model incapable of easily explaining the autonomy
effects demonstrated by Forster (1979), Frazier (1987), and others.
As shown above, COMPERE's unified processing mechanism combined with
its separate sources of linguistic knowledge offers an explanation for
observed autonomy effects as well as the interaction effects reported
by Marslen-Wilson and Tyler (Tyler \& Marslen-Wilson, 1977).
Furthermore, the integrated models noted above cannot capture
syntactic generalizations.

Another form of the modularity debate concerns the effect of context
on syntactic decisions---does context affect structure assignment, or
are context effects absent until later in language processing (Taraban
\& McClelland, 1985)?  Though we do not have a model of context
effects in COMPERE, we believe that contextual information can be
incorporated as an additional source of preferences in COMPERE's
architecture.

An added benefit of COMPERE's sentence processing architecture is that
it offers an explanation for the effects of linguistic aphasias.  In
reviewing the aphasia literature, Caramazza and Berndt (1978)
concluded that the evidence pointed strongly to the functional
independence of syntactic and semantic processing.  COMPERE suggests
an alternate explanation---the different aphasic behaviors are not due
to damage to the individual processors, but are instead due to damage
to the individual knowledge sources or, perhaps, to the communications
pathways between the knowledge sources and the unified processor.

We believe that COMPERE's architecture accounts for the wide variety
of seemingly conflicting data on linguistic behavior better than any
previously proposed model of sentence processing.  Yet COMPERE is not
the first sentence processing model to be configured as a single
process interacting with independent knowledge sources.  The localist
or punctate connectionist models of Pollack (1987; Waltz and Pollack,
1985) and Cottrell (1985; Cottrell and Small, 1983) resemble COMPERE
at a gross architectural level, but these models did not offer the
range of explanation of different behaviors that COMPERE does; for
example, these models do not recover from errors, nor can they deal
with complex syntactic structures such as relative clauses.

Despite all its theoretical advantages over other models, the
prototype implementation of COMPERE is not yet fully developed and
suffers from some weaknesses.  Its role knowledge is fairly limited,
and its conceptual knowledge is even more so.  Also, the
implementation currently diverges slightly from theory.  The
divergence appears in the process itself: the theoretical model has a
single unified process, while the prototype computational model
consists of two nearly-identical processes---one for syntax and one
for semantics.  These two processes share identical control
structures, but they are duplicated because we have not yet completed
the task of representing the different types of information in a
uniform format.  Some readers may take this as an indication that we
are doomed to failure, but the connectionist models mentioned earlier
serve as existence proofs that finding a uniform format for
representing different types of linguistic knowledge is by no means an
impossible task.

\section{Conclusion}

Is the human language understander a collection of modular processes
operating with relative autonomy, or is it a single integrated
process?  This ongoing debate has polarized the language processing
community, with two fundamentally different types of model posited,
and with each camp concluding that the other is wrong.  One camp puts
forth a model with separate processors and distinct knowledge sources
to explain one body of data, and the other proposes a model with a
single processor and a homogeneous, monolithic knowledge source to
explain the other body of data.  In this paper we have argued that a
hybrid approach which combines a unified processor with separate
knowledge sources provides an explanation of both bodies of data, and
we have demonstrated the feasibility of this approach with the
computational model called COMPERE.  We believe that this approach
brings the language processing community significantly closer to
offering human-like language processing systems.

\smallskip
\noindent{\bf Acknowledgement:} We would like to thank Justin Peterson 
for his comments on this work and his help in finding good examples.

\section{References}
\noindent{}Birnbaum, L., and Selfridge, M. 1981.  Conceptual analysis of 
natural language.  In Schank, R. C., and Riesbeck, C. K. eds. Inside
computer understanding: Five programs plus miniatures, 318-353.
Hillsdale, NJ: Lawrence Erlbaum.

\vskip 0.04 in
\noindent{}Caramazza, A., and Berndt, R. S. 1978.  Semantic and syntactic 
processes in aphasia: A review of the literature.  {\em Psychological
Bulletin} 85:898-918.

\vskip 0.04 in
\noindent{}Cottrell, G. W. 1985.  A connectionist approach to word sense 
disambiguation, Technical Report, 154, Computer Science Department,
University of Rochester.

\vskip 0.04 in
\noindent{}Cottrell, G. W., and Small, S. L. 1983.  A connectionist scheme 
for modelling word sense disambiguation.  {\em Cognition and Brain
Theory} 6:89-120.

\vskip 0.04 in
\noindent{}Crain, S., and Steedman, M. 1985.  On not being led up the 
garden path: The use of context by the psychological syntax processor.
In Dowty, D. R., Kartunnen, L., and Zwicky, A. M. eds. Natural
language parsing: Psychological, computational, and theoretical
perspectives, 320-358.  Cambridge, England: Cambridge University
Press.

\vskip 0.04 in
\noindent{}Eiselt, K. P. 1987.  Recovering from erroneous inferences.  In 
Proc. AAAI-87 Sixth National Conference on Artificial Intelligence,
540-544.  San Mateo, CA: Morgan Kaufmann.

\vskip 0.04 in
\noindent{}Eiselt, K. P. 1989.  Inference processing and error recovery in 
sentence understanding, Technical Report, 89-24, Ph.D. diss., Dept. of
Computer Science, University of California, Irvine.

\vskip 0.04 in
\noindent{}Eiselt, K. P., and Holbrook, J. K. 1991.  Toward a unified theory 
of lexical error recovery.  In Proc. of the Thirteenth Annual
Conference of the Cognitive Science Society, 239-244.  Hillsdale, NJ:
Lawrence Erlbaum.

\vskip 0.04 in
\noindent{}Forster, K. I. 1979.  Levels of processing and the structure of 
the language processor.  In Cooper, W. E., and Walker,
E. C. T. eds. Sentence processing: Psycholinguistic studies presented
to Merrill Garrett, 27-85.  Hillsdale, NJ: Lawrence Erlbaum.

\vskip 0.04 in
\noindent{}Frazier, L. 1987.  Theories of sentence processing.  In 
Garfield, J. L. ed.  Modularity in knowledge representation and
natural-language understanding.  Cambridge, MA: MIT Press.

\vskip 0.04 in
\noindent{}Granger, R. H., Eiselt, K. P., and Holbrook, J. K. 1983.  
STRATEGIST: A program that models strategy-driven and content-driven
inference behavior. In Proc.  of the National Conference on Artificial
Intelligence, 139-147.  San Mateo, CA: Morgan Kaufmann.

\vskip 0.04 in
\noindent{}Holbrook, J. K. 1989.  Studies of inference retention in lexical 
ambiguity resolution.  Ph.D. diss., School of Social Sciences,
University of California, Irvine.

\vskip 0.04 in
\noindent{}Holbrook, J. K., Eiselt, K. P., and Mahesh, K. 1992.  A unified 
process model of syntactic and semantic error recovery in sentence
understanding. In Proc. of the Fourteenth Annual Conference of the
Cognitive Science Society, 195-200.  Hillsdale, NJ: Lawrence Erlbaum.

\vskip 0.04 in
\noindent{}Jurafsky, D. 1992.  An on-line computational model of human 
sentence interpretation.  In Proc. of the Tenth National Conference on
Artificial Intelligence, 302-308.  San Mateo, CA: Morgan Kaufmann.

\vskip 0.04 in
\noindent{}Kurtzman, H. S. 1985.  Studies in syntactic ambiguity resolution.  
Ph.D. diss., Dept. of Psychology, Massachusetts Institute of
Technology.

\vskip 0.04 in
\noindent{}Lebowitz, M. 1980.  Generalization and memory in an integrated 
understanding system, Research Report, 186, Dept. of Computer Science,
Yale University.

\vskip 0.04 in
\noindent{}Lytinen, S. L. 1984.  The organization of knowledge in a 
multi-lingual, integrated parser, Research Report, YALEU/CSD/RR 340,
Dept. of Computer Science, Yale University.

\vskip 0.04 in
\noindent{}Marcus, M. P. 1980.  {\em A theory of syntactic recognition for 
natural language.}  Cambridge, MA: MIT Press.

\vskip 0.04 in
\noindent{}Pollack, J. B. 1987.  On connectionist models of natural language 
processing, Technical Report, MCCS-87-100, Computing Research
Laboratory, New Mexico State University.

\vskip 0.04 in
\noindent{}Riesbeck, C. K., and Martin, C. E. 1986.  Towards completely 
integrated parsing and inferencing.  In Proc. of the Eighth Annual
Conference of the Cognitive Science Society, 381-387.  Hillsdale, NJ:
Lawrence Erlbaum.

\vskip 0.04 in
\noindent{}Seidenberg, M. S., Tanenhaus, M. K., Leiman, J. M., and 
Bienkowski, M. 1982.  Automatic access of the meanings of ambiguous
words in context: Some limitations of knowledge-based processing.
{\em Cognitive Psychology} 14:489-537.

\vskip 0.04 in
\noindent{}Stowe, L. A. 1991.  Ambiguity resolution:  Behavioral evidence for 
a delay.  In Proc. of the Thirteenth Annual Conference of the
Cognitive Science Society, 257-262.  Hillsdale, NJ: Lawrence Erlbaum.

\vskip 0.04 in
\noindent{}Taraban, R., and McClelland, J. L. 1988.  Constituent attachment 
and thematic role assignment in sentence processing: Influences of
content-based expectations.  {\em Journal of Memory and Language}
27:597-632.

\vskip 0.04 in
\noindent{}Tyler, L. K., and Marslen-Wilson, W. D. 1977.  The on-line 
effects of semantic context on syntactic processing.  {\em Journal of
Verbal Learning and Verbal Behavior} 16:683-692.

\vskip 0.04 in
\noindent{}Waltz, D. L., and Pollack, J. B. 1985. Massively parallel 
parsing: A strongly interactive model of natural language
interpretation.  {\em Cognitive Science} 9:51-74.

\vskip 0.04 in
\noindent{}Woods, W. A. 1973. An experimental parsing system for 
transition network grammars.  In Rustin, R. ed. Natural language
processing.  New York: Algorithmics Press.

\end{document}